\documentclass[a4paper]{spie}  

\usepackage{amsmath,amsfonts,amssymb}
\usepackage{graphicx}
\usepackage[colorlinks=true, allcolors=blue]{hyperref}

\usepackage{booktabs}

\usepackage{tabularx}
\newcolumntype{C}{>{\centering\arraybackslash}X}

\usepackage{multirow}
\usepackage{amssymb}
\usepackage{textcomp}

\def\vec#1{\ensuremath{\mathchoice
                     {\mbox{\boldmath$\displaystyle#1$}}
                     {\mbox{\boldmath$\textstyle#1$}}
                     {\mbox{\boldmath$\scriptstyle#1$}}
                     {\mbox{\boldmath$\scriptscriptstyle#1$}}}}

\usepackage{tikz}
\usetikzlibrary{external}
\title{Semantic denoising autoencoders for retinal optical coherence tomography}

\author{Max-Heinrich Laves}
\author{Sontje Ihler}
\author{L{\"u}der Alexander Kahrs}
\author{Tobias Ortmaier}
\affil{Institute of Mechatronic Systems, Leibniz Universit{\"a}t Hannover,\\Appelstr. 11A, 30167 Hannover, Germany}

\authorinfo{Send correspondence to:\\
M.H. Laves: E-mail: laves@imes.uni-hannover.de, Telephone: +49\,511\,762\,19617}

\begin{document} 
\maketitle

\begin{abstract}
Noise in speckle-prone optical coherence tomography tends to obfuscate important details necessary for medical diagnosis.
In this paper, a denoising approach that preserves disease characteristics on retinal optical coherence tomography images in ophthalmology is presented.
By combining a deep convolutional autoencoder with a priorly trained ResNet image classifier as regularizer, the perceptibility of delicate details is encouraged and only information-less background noise is filtered out.
With our approach, higher peak signal-to-noise ratios with $ \mathrm{PSNR} = 31.2\,\mathrm{dB} $ and higher classification accuracy of $\mathrm{ACC} = 85.0\,\%$ can be achieved for denoised images compared to state-of-the-art denoising with $ \mathrm{PSNR} = 29.4\,\mathrm{dB} $ or $\mathrm{ACC} = 70.3\,\%$, depending on the method.
It is shown that regularized autoencoders are capable of denoising retinal OCT images without blurring details of diseases.
\end{abstract}


\section{Purpose}
\label{sec:intro}

Optical coherence tomography (OCT) is the most common imaging technique for diagnosis in ophthalmology.
However, due to image acquisition based on interference of coherent light, OCT suffers from speckle noise.
This results in grainy images with low contrast where the diagnosis of medical conditions requires trained expert observers.
Denoising of OCT has been addressed in the literature already and can be separated into two categories\cite{Salinas2007}.
The first one employs denoising during OCT acquisition by e.g.\ averaging multiple frames of the same object.
This prolongs the acquisition process and is therefore not applicable for dynamic objects.
The second category comprises post-processing methods such as median, bilateral, wavelet-based or other linear and nonlinear filtering techniques.
These can be executed in real time but are prone not only to blurring the image, but also to erasing important disease-related details in that process.
This paper describes a domain-specific post-processing method for denoising OCT images with machine learning and more specific convolutional autoencoders (AE) while maintaining disease characteristics.

\section{Methods}

The dataset used in this paper contains 84,484 retinal OCT images from 4,657 patients showing the disease states drusen, diabetic macular edema (DME), choroidal neovascularization (CNV) and normal and is publicly available\cite{kermany2018}.
First, a ResNet-34 image classifier $ C $ pretrained on ImageNet is fine-tuned on the dataset\cite{He2016}.
This acts as medical expert as it has been shown that the performance of convolutional neural networks (CNNs) in classifying retinal conditions is on par to that of trained ophthalmologists \cite{kermany2018}.
Second, the ErfNet CNN autoencoder is trained to reconstruct input images $ \vec{x} $ corrupted by additive gaussian white noise resulting in $ \vec{\tilde{x}} = \vec{x} + \vec{c} $ with $ \vec{c} \sim \mathcal{N}(\vec{0}, 0.1\,\mathbf{I}) $\cite{Romera2018}.
In general, an AE consists of two components.
The encoder $ E $ takes an input image $ \vec{x} $, or in our case $ \vec{\tilde{x}} $, and maps it from high dimension into low-dimensional, latent representation $ \vec{z} $.
This is then fed into the decoder $ D $ and mapped back to a reconstructed image $ \hat{\vec{x}} $ in input space.
The parameters of the AE are optimized by minimizing the pixel-wise mean squared reconstruction error $ L_{r}(\vec{x}, \hat{\vec{x}}) $.
Essentially, an autoencoder learns a low-dimensional representation similar to principal component analysis (PCA).
When training with a large dataset, noise tends to ``average out'' and the AE reconstructs distinct and relevant (noise-free) image features.
In order to promote enhancement of these features, the trained ResNet with fixed weights is used as additional optimization criterion $ L_c $.
It is applied to the reconstructed, denoised image and tries to predict the retinal disease class.
This regularizes the AE during training and enhances disease characteristics in denoised images.
The proposed approach is therefore optimized using the weighted loss function
\begin{equation}
	L_{\mathrm{AE}}(\vec{x}, \hat{\vec{x}}, \vec{y}) = L_r(\vec{x}, \hat{\vec{x}}) + \alpha L_c(\vec{y}, C(\vec{x}))
\end{equation}
with denoised corrupted image $ \hat{\vec{x}} = D(E(\tilde{\vec{x}})) $, true disease label $ \vec{y} $ of image $ \vec{x} $ and cross entropy for $ L_c $.
Weighting factor for $ L_c $ was empirically set to $ \alpha = 0.1 $.
The aforementioned method is implemented with PyTorch 1.0 and trained for 200 epochs using the Adam optimizer with an initial learning rate of $ \eta = 10^{-4} $\cite{Kingma2014}.
A reduce-on-plateau learning rate scheduling is realized to reduce $ \eta $ with a factor of $ 10^{-1} $ when observing saturation of the validation loss.
The weight configuration with lowest loss value on the validation set is chosen for testing (early stopping).

\section{Results}
\label{sec:results}

The CNN are optimized using 79,484 OCT images for training, 4,000 for validation and 1,000 for testing.
To assess denoising performance, the proposed method is compared to total variation (TV) minimization\cite{Chambolle2004}, wavelet\cite{Chang2000}, and anisotropic diffusion (AD)\cite{Perona1990} denoising regarding peak signal-to-noise ratio (PSNR) and classification performance of ResNet.
\begin{table}[tb]
  \centering
  \begin{tabular}{lccccc}
    \toprule
     & corrupted & TV & wavelet    & AD & AE (ours) \\
    \cmidrule(lr){2-6}
    PSNR & 19.2 & 29.4 & 28.0 & 24.6 & \textbf{31.2} \\
    ACC  & 50.2 & 49.3 & 52.6 & 70.3 & \textbf{85.0} \\
    \bottomrule \\
  \end{tabular}
  \caption{Results of denoising reported for test set with mean peak signal-to-noise ratio PSNR in dB and mean classification accuracy ACC in \%. Values for corrupted images $ \tilde{\vec{x}} $ are given for comparison. Bold values denote best results.}
  \label{tab:results}
\end{table}
The results are summarized in Tab.~\ref{tab:results}.
Our approach not only provides the highest disease classification accuracy with $ \mathrm{ACC} = 85.0\,\% $ after denoising, but also has the highest peak signal-to-noise ration with $ \mathrm{PSNR} = 31.2\,\mathrm{dB} $ compared to the other methods.
\begin{figure}
	\newcommand{\imgWidth}{2.6cm}
	\newcommand{\imgWidthHalf}{1.3cm}
	\setlength\tabcolsep{2pt}
	\renewcommand{\arraystretch}{0}
    \centering
    \small
    \begin{tabular}{ccccccc}
    & $ \vec{x} $ (original) & $ \tilde{\vec{x}} $ (corrupted) & total variation & wavelet & AD & AE (ours) \\ \addlinespace[0.25em]
    \raisebox{\imgWidthHalf}{\rotatebox[origin=c]{90}{NORMAL}} &
    \includegraphics[width=\imgWidth]{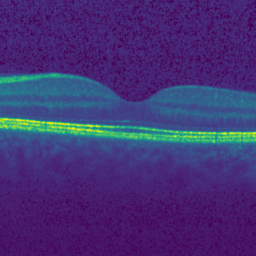} &
    \includegraphics[width=\imgWidth]{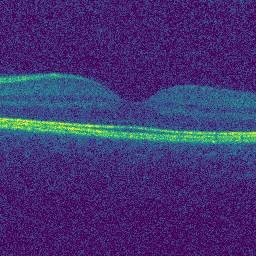} &
    \includegraphics[width=\imgWidth]{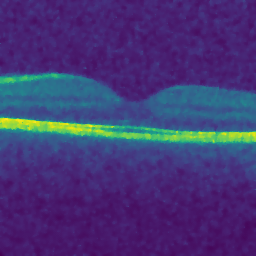} &
    \includegraphics[width=\imgWidth]{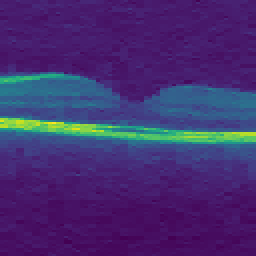} &
    \includegraphics[width=\imgWidth]{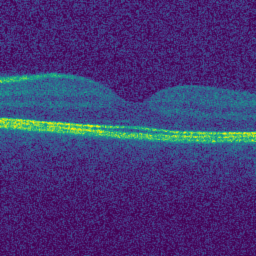} &
    \includegraphics[width=\imgWidth]{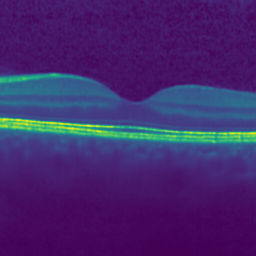} \\ \addlinespace[0.5em]
\raisebox{\imgWidthHalf}{\rotatebox[origin=c]{90}{DRUSEN}} &
    \includegraphics[width=\imgWidth]{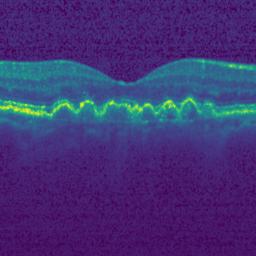} &
    \includegraphics[width=\imgWidth]{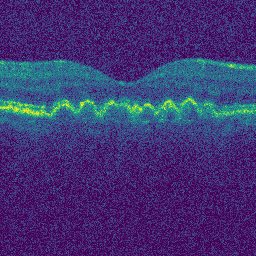} &
    \includegraphics[width=\imgWidth]{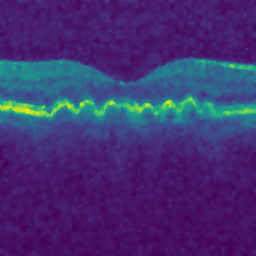} &
    \includegraphics[width=\imgWidth]{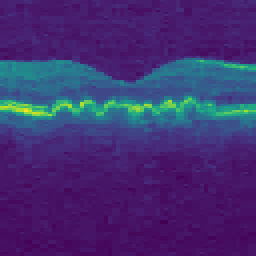} &
    \includegraphics[width=\imgWidth]{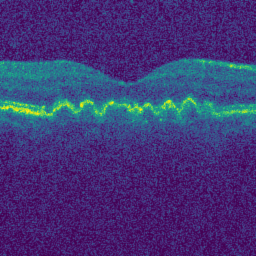} &
    \includegraphics[width=\imgWidth]{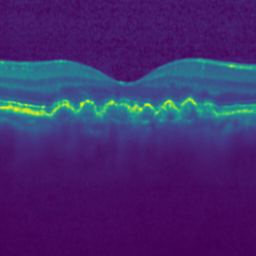} \\ \addlinespace[0.5em]
\raisebox{\imgWidthHalf}{\rotatebox[origin=c]{90}{DME}} &
    \includegraphics[width=\imgWidth]{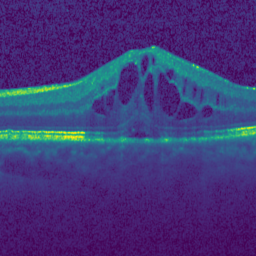} &
    \includegraphics[width=\imgWidth]{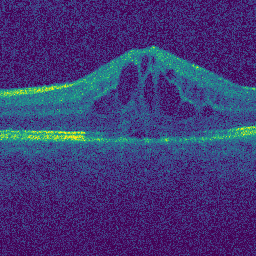} &
    \includegraphics[width=\imgWidth]{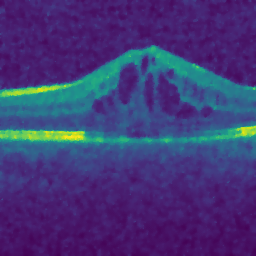} &
    \includegraphics[width=\imgWidth]{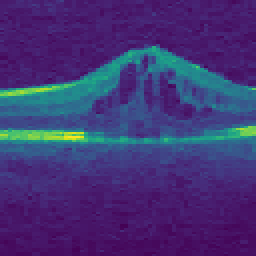} &
    \includegraphics[width=\imgWidth]{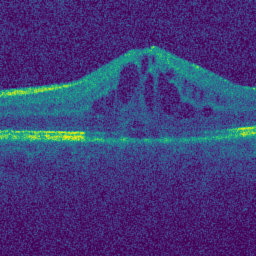} &
    \includegraphics[width=\imgWidth]{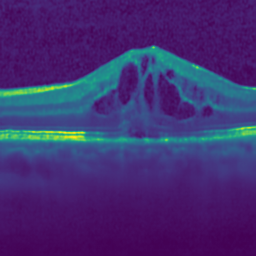} \\ \addlinespace[0.5em]
\raisebox{\imgWidthHalf}{\rotatebox[origin=c]{90}{CNV}} &
    \includegraphics[width=\imgWidth]{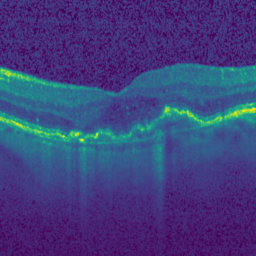} &
    \includegraphics[width=\imgWidth]{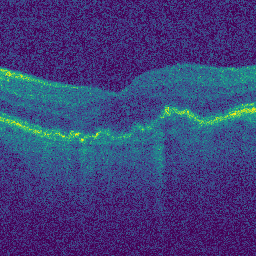} &
    \includegraphics[width=\imgWidth]{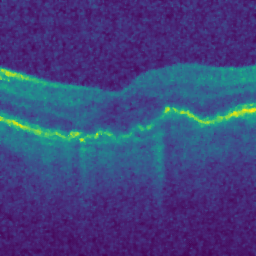} &
    \includegraphics[width=\imgWidth]{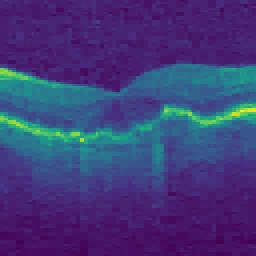} &
    \includegraphics[width=\imgWidth]{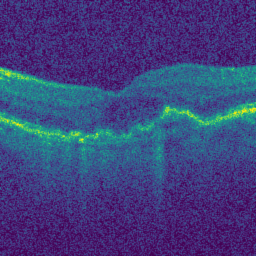} &
    \includegraphics[width=\imgWidth]{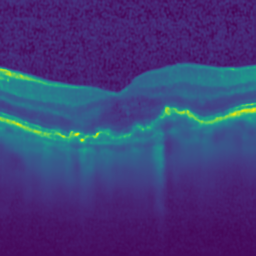} \\ \addlinespace[0.5em]
    \end{tabular}
  \caption{Results of our approach compared to state-of-the-art denoising for retinal OCT disease conditions from the test set. Digital zoom is recommended for optimal comparison.}
  \label{fig:results}
\end{figure}
Fig.~\ref{fig:results} visualizes qualitative results for example OCT from the test set showing different disease conditions.
The methods are used to restore the input image $ \vec{x} $ (first column) from the corrupted image $ \tilde{\vec{x}} $ (second column).
In contrast to state-of-the-art denoising, our approach is able to distinctively preserve the retinal layers while removing speckle noise.
Pathological alterations of the retina are clearly visible and the  explanatory power for diagnosis is not reduced.
Mean processing time of AE for one image is 13.1\,ms on an NVIDIA GeForce GTX 1080\,Ti.

\section{Conclusion}
\label{sec:conclusion}

It has been shown that convolutional AEs are capable of denoising retinal OCT images without suppressing characteristics of diseases.
This was achieved by regularizing the denoising AE during training with another CNN, which was previously trained for disease classification.
The trained decoder can also be used to generate new images by sampling the latent space.
Future work therefore aims on variational AEs and generative adversarial networks for OCT denoising.
It should be noted, however, that speckle noise can also contain significant information as it creates a unique fingerprint of tissue.
This information cannot be interpreted by humans, and CNNs can be valuable tools to acquire and utilize this information in the future.

\paragraph{Conflict of Interest} The authors declare that they have no conflict of interest.


\paragraph{Formal Consent} The medical images used in this article were made available to the public in a previous study \cite{kermany2018}, therefore formal consent is not required.

%

\bibliography{literature} 
\bibliographystyle{spiebib} 

\end{document}